\documentclass[10pt]{article}

\usepackage{algorithm}
\usepackage{algpseudocode}
\usepackage{amsmath}
\usepackage{amsfonts}  
\usepackage{booktabs}
\usepackage{multirow}
\usepackage{ulem}
\usepackage{xcolor}
\usepackage{graphicx}
\usepackage{subcaption}
\usepackage{geometry}
\usepackage{hyperref} 
\usepackage[numbers]{natbib}  
\newcommand{\ForParallel}[1]{\For{each client $#1$ \textbf{in parallel}}}

\title{	FedAli: Personalized Federated Learning Alignment with Prototype Layers for Generalized Mobile Services}

\author{
    Sannara Ek \\
    Université Grenoble Alpes \\
    Grenoble, France \\
    \texttt{sannara.ek@gmail.com}
    \and
    Kaile Wang \\
    Hong Kong Polytechnic University \\
    Hong Kong (SAR), China \\
    \texttt{kacy1.wang@connect.polyu.hk}
    \and
    François Portet \\
    Université Grenoble Alpes \\
    Grenoble, France \\
    \texttt{francois.portet@imag.fr}
    \and
    Philippe Lalanda \\
    Université Grenoble Alpes \\
    Grenoble, France \\
    \texttt{philippe.lalanda@imag.fr}
    \and
    Jiannong Cao \\
    Hong Kong Polytechnic University \\
    Hong Kong (SAR), China \\
    \texttt{csjcao@comp.polyu.edu.hk}
}

\date{}

\begin{document}

\maketitle

\begin{abstract}

Personalized Federated Learning (PFL) enables distributed training on edge devices, allowing models to collaboratively learn global patterns while tailoring their parameters to better fit each client's local data, all while preserving data privacy. However, PFL faces two key challenges in mobile systems: client drift, where heterogeneous data cause model divergence, and the overlooked need for client generalization, as the dynamic of mobile sensing demands adaptation beyond local environments. To overcome these limitations, we introduce Federated Alignment (FedAli), a prototype-based regularization technique that enhances inter-client alignment while strengthening the robustness of personalized adaptations. At its core, FedAli introduces the ALignment with Prototypes (ALP) layer, inspired by human memory, to enhance generalization by guiding inference embeddings toward personalized prototypes while reducing client drift through alignment with shared prototypes during training. By leveraging an optimal transport plan to compute prototype-embedding assignments, our approach allows pre-training the prototypes without any class labels to further accelerate convergence and improve performance. Our extensive experiments show that FedAli significantly enhances client generalization while preserving strong personalization in heterogeneous settings.

\end{abstract}



\section{Introduction} \label{sec:intro}


Advancements in sensor technology, memory, and computational capabilities in everyday edge devices, such as smartphones and smartwatches, have significantly expanded the scope of on-device AI services \cite{laskaridis2024future,li2025survey}. However, these systems still face major challenges in mobile pervasive computing, particularly the strong heterogeneity of clients and the continuous evolution of data, which undergo various types of shifts. This variability, which leads to data that is non-Independent and Identically Distributed (Non-IID), is further exacerbated in real-world environments, where data distributions vary significantly among users due to individual habits, local conditions, and other contextual factors \citep{10.5555/1462129,wang2022generalizing,lu2022semantic}. 


Federated Learning (FL) tackles critical challenges by enabling model training directly on edge devices, eliminating the need for centralized data collection. This decentralized paradigm significantly enhances data security as sensitive information remains on user devices, thereby reducing privacy risks. In addition, FL optimizes the use of computing resources by training models on a diverse array of real-world datasets distributed across various users. Nevertheless, conventional FL still struggles with client heterogeneity, since a single global model may not achieve optimal performance for every user \citep{mcmahan2017communication,li2020federated,li2020federatedOpt,kairouz2021advances,li2021survey,liao2023adaptive}.


To address this challenge, Personalized Federated Learning (PFL) extends FL by tailoring models to individual clients \citep{zhang2021personalized,tan2022towards}. Rather than enforcing a uniform global model, PFL adapts training strategies to better fit each user's specific data distribution. Techniques such as local fine-tuning and meta-learning enable PFL to strike a balance between generalization and personalization \citep{tan2022towards}. However, most PFL approaches \citep{li2021ditto,tan2022fedproto,mu2023fedproc,xu2023personalized} have been designed and evaluated exclusively on the client's local data, often neglecting the model's ability to generalize beyond the environment of each user.

In this study, we argue that high-quality personalized models must satisfy two essential properties: (1) the ability to adapt effectively to local data (personalization) and (2) robustness to unseen scenarios (generalization). Concretely, we reinterpret the weight-sharing mechanism of FL as a form of domain transfer learning \citep{qin2019cross,chen2020fedhealth,nguyen2022fedsr,guo2023out}, using shared knowledge between clients to enhance the generalization capacity of personalized models.

For instance, in the Wearable Human Activity Recognition (HAR) domain, mobile sensing devices such as smartphones and smartwatches operate in dynamic environments and must continuously adapt to evolving user behaviors \citep{mukhopadhyay2014wearable,xu2023practically}. Ideally, a model trained collaboratively through PFL for a user in an urban setting should accurately detect activities on flat surfaces. In addition, it must also remain effective when the same user transitions to a mountainous environment with rough surfaces that can affect the sensor's reading. Achieving this level of robustness requires leveraging knowledge shared by other users who are in different environments and conditions. By utilizing the FL framework, the model can enhance its generalization beyond the client's local environment, ensuring robustness across varying contexts.

To address the challenges posed by data heterogeneity in FL (ensuring alignment across clients) and to enhance the robustness of client models (enabling them to perform well on locally unseen data), we introduce a novel neural network layer: ALignment with Prototypes (ALP). Inspired by human memory, the ALP layer influences both training and inference by leveraging trained prototypes to guide the client's model. At the core of our approach is a new PFL strategy, Federated Alignment (FedAli), which coordinates the ALP layer to align and regularize client model embeddings, improving generalization across diverse conditions.

The key contributions of this work are as follows:
\begin{itemize}
\item We highlight the importance of client model generalization in PFL works applied to mobile computing systems and analyze its impact under various FL strategies.
\item We introduce FedAli, our novel PFL strategy that leverages prototypes through our proposed ALP layer. This layer enhances the alignment and generalization of distributed learning models between clients utilizing an embedding-to-prototypes matching strategy based on an optimal transport plan.
\item Under rigorous evaluation on heterogeneous HAR and benchmark vision datasets, we demonstrate that FedAli outperforms other approaches in client model generalization while maintaining strong personalization.
\end{itemize}

The paper is structured as follows: Section 2 formalizes our learning objective and related PFL work. Section 3 introduces our key concept, while Section 4 details the proposed layer and FL strategy. Section 5 presents experimental results, and Section 6 provides an ablation analysis. Finally, Section 7 concludes the work.

\section{Background and related work} \label{sec:preliminsries}



The conventional FL framework consists of multiple clients collaboratively training a global model without ever communicating client data \citep{mcmahan2017communication}. The baseline FedAvg algorithm defines the learning objective as below:

\begin{equation}\label{eq:FL_eq}
     \underset{w}{min} F(w;\mathcal{D}) =\sum_{c=1}^{C}\dfrac{n_c}{n} f_c(w;\mathcal{D}_c) 
\end{equation}

where $F(.)$ is the FL objective, $f_{c}(.)$ the local loss of client $c$, $w$ the weights of the global model, $C$ the number of clients, $n$ the total count of the data samples and $n_{c}$ the number of data samples of local client $c$, $\mathcal{D}_c$ the local dataset of client $c$ and $\mathcal{D}$ the combined data sets of all clients. We note here that $\mathcal{D}$ is actually never available for the global model at the server level in practice (except in simulations to showcase the learning performance of the global model). 

However, traditional FL follows a server-centric approach, primarily aimed at optimizing the performance of a global model. In this framework, relying on a single global model may prove insufficient to effectively address variations among client models, potentially overlooking the unique characteristics and data distributions of individual users.

Personalized federated learning is an advanced extension of FL that aims to tailor the learning process to the unique requirements and data characteristics of individual clients \citep{tan2022towards}. Recalling Equation \ref{eq:FL_eq}, our interpretation of the learning objective of PFL can be expressed as follows:



\begin{equation}\label{eq:PerFL_eq}
\begin{aligned}
 F(w;\mathcal{D}) &= \sum_{c=1}^{C} \dfrac{n_c}{n} f_c(w;\mathcal{D}_c), \quad
 \text{where} \quad \forall c \quad \underset{w_c}{\min} f_c(w_c;\mathcal{D}_c) \\
 &\hspace{2cm} \text{s.t.} \quad H(w_c;\mathcal{D}) \text{ is minimized}
\end{aligned}
\end{equation}

where $H(\cdot)$ an error measurement on the entire dataset of combined clients $\mathcal{D}$. The change in expression signifies a user-centric approach, where the main goal is now to have each client $c$ optimize the model $w_c$ on their local data set $\mathcal{D}_c$ to increase personalization \citep{zhang2021personalized}. Furthermore, we argue that the evaluation function $H(\cdot)$ is needed to measure how optimally the local model $w_c$ performs on the global dataset $\mathcal{D}$ to estimate the generalization performances. 





Research in PFL has aimed to mitigate the challenges posed by heterogeneity through various regularization techniques. A widely adopted approach involves leveraging contrastive learning \citep{chen2020simple} or distillation techniques \citep{hinton2015distilling} to penalize dissimilarity, thereby facilitating knowledge transfer from a shared source to local models during training. Several studies have explored different sources for this knowledge transfer: some propose extracting embeddings from intermediate client layers \citep{collins2021exploiting, liang2020think, 10286439, yang2024fedfed}, while others rely on the global model or leverage models from previous clients \citep{acar2021debiasing, li2021model, li2019fedmd, li2021ditto, chen2023spectral}.



More recently, class prototypes have been introduced as a means to regularize similarity across clients \citep{snell2017prototypical}. This approach involves generating a set of prototype vectors, typically the barycenters of each class's embeddings across all clients, to provide a consistent representation of the data throughout the network. Recent studies have used these prototypes as a form of global knowledge to better align local client training with the global goals underlying \citep{michieli2021prototype, dai2023tackling, xu2023personalized, mu2023fedproc}. By mitigating inconsistencies in model training between clients, these methods seek to enhance collaboration despite data heterogeneity.

However, class prototype-based regularization has notable limitations. First, it inherently relies on class labels, restricting its applicability to supervised learning scenarios. Additionally, these approaches typically assign a single prototype per class, which may be insufficient in cases where a class exhibits strong domain-induced variations. This oversimplification can negatively impact performance in distributed settings, where the same class can manifest differently across client domains.

Finally, as previously mentioned, these approaches often neglect the generalization performance of client models. We argue that clients could further benefit from adapting unseen local data at inference time to better align with the learned prototypes from training. This adaptation has the potential to enhance model consistency and robustness, particularly in heterogeneous environments where local data distributions vary significantly.


\section{Proposed Approach} \label{sec:method}

\begin{figure}[ht]
\centering
\includegraphics[width=0.6\linewidth]{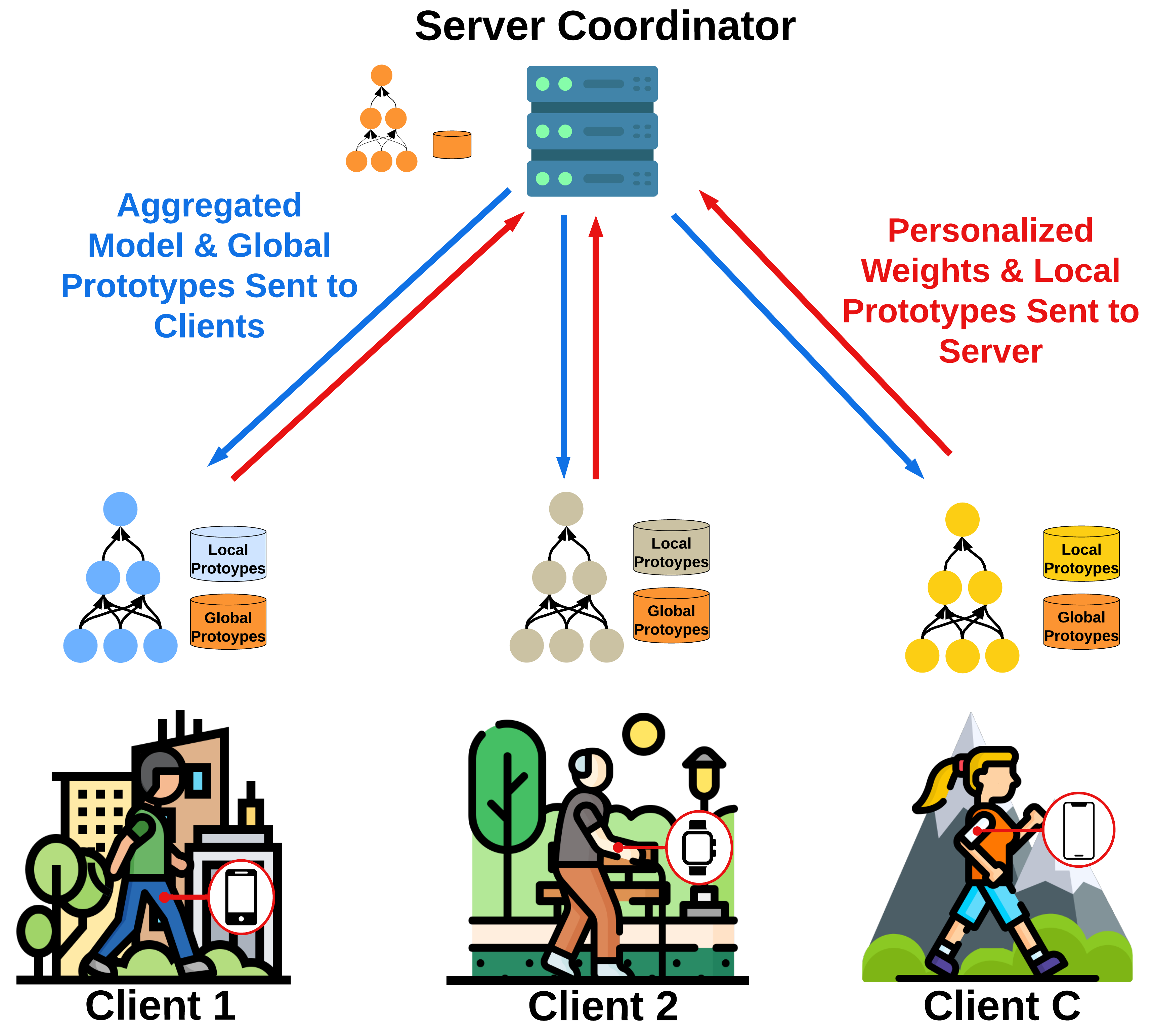}
\caption{Federated Alignment with heterogeneous clients}
\label{Fig:fedAli}
\end{figure}

\subsection{Overview}

To mitigate client drift and enhance the robustness of personalized client models, we introduce FedAli, as illustrated in Figure \ref{Fig:fedAli}. FedAli leverages a novel prototype-based regularization technique that operates without requiring class labels, using ALP layers to improve model alignment and generalization. Inspired by how human memory processes new stimuli \citep{barsalou2009simulation}, the ALP layer is designed to relate new information to prior knowledge, akin to how conceptual representations are built from past experiences and simulations (training data).

A prototype is a representative vector that encapsulates essential features of learned data, serving as a structured memory unit for comparison and alignment. In FedAli, we employ personalized local prototypes, which act as the memory of individual clients, ensuring that new inputs align with past experiences during inference, thus enhancing model robustness. Additionally, we introduce shared global prototypes, which serve as common reference points across clients, encouraging consistency in embeddings, and promoting similarity in learned representations. This novel approach to client regularization reduces divergence in training data distributions, ultimately leading to more aligned and generalizable models across clients.

We further elaborate on the concept of these two types of prototypes below.




\subsection{Local prototypes: inference alignments}

As introduced, we use banks of trainable prototypes as personalized memory representations of each client's previously acquired knowledge. Within the ALP layer, these prototypes are optimally matched to their corresponding embeddings through a balanced optimal transport plan \citep{villani2009optimal,peyre2019computational}. During local training, prototype sets are continuously updated toward their assigned embeddings within each data batch using an Exponential Moving Average (EMA), effectively "remembering" learned information. We refer to this evolving set of prototypes, which captures and retains representative concepts or features from local data, as the local prototypes.

During model inference, these personalized local prototypes are matched with the embeddings, ensuring that all input data (whether previously encountered or novel) align more closely with the past training characteristics. By transforming the embeddings to better fit the model’s learned structure, our approach significantly enhances the robustness of local models.

\subsection{Global prototypes: training alignments}

While each client maintains a unique set of local prototypes, introducing additional shared prototypes can foster collaboration within the FL framework.

Since local prototypes function as trainable weights, they are transmitted alongside the trained client weights during the FL aggregation stage. The FL server then consolidates these local prototypes into a fixed number of representative prototypes, which we refer to as global prototypes.

Unlike previous approaches that assign a single barycentric prototype per class as the global prototype \citep{li2021model,tan2022fedproto,mu2023fedproc}, our method operates without requiring class labels and leverages a larger set of prototypes to better accommodate domain heterogeneity across clients. Instead of relying on predefined class associations, we compute global prototypes using clustering techniques, ensuring that they capture diverse patterns across different local distributions. The resulting global prototypes are then distributed back to clients in the next communication round.

These global prototypes serve as embedding anchors \citep{zhou2023fedfa}, guiding feature embeddings toward a shared representation across all clients during training. Unlike previous methods \citep{li2021model,tan2022fedproto,mu2023fedproc}, which rely on auxiliary loss terms to enforce consistency (while still allowing divergence in feature extractor weights), our novel client regularization approach enhances consistency by directly aligning training data embeddings. This results in more stable and coherent model representations across clients, effectively mitigating representation drift, and improving overall model robustness.

\section{Design \& Implementation} \label{sec:implement}

\begin{figure}[ht]
\centering
\includegraphics[width=0.35\linewidth]{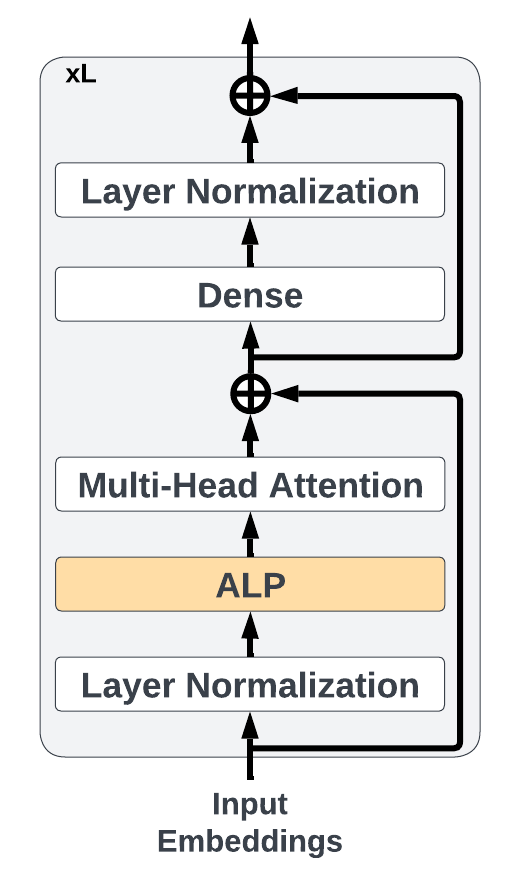}
\caption{The ALP layer is positioned after the first normalization layer within the transformer encoder block.}
\label{Fig:ALP_position}
\end{figure}


\begin{figure*}[ht]
\centering
\includegraphics[width=1.0\linewidth]{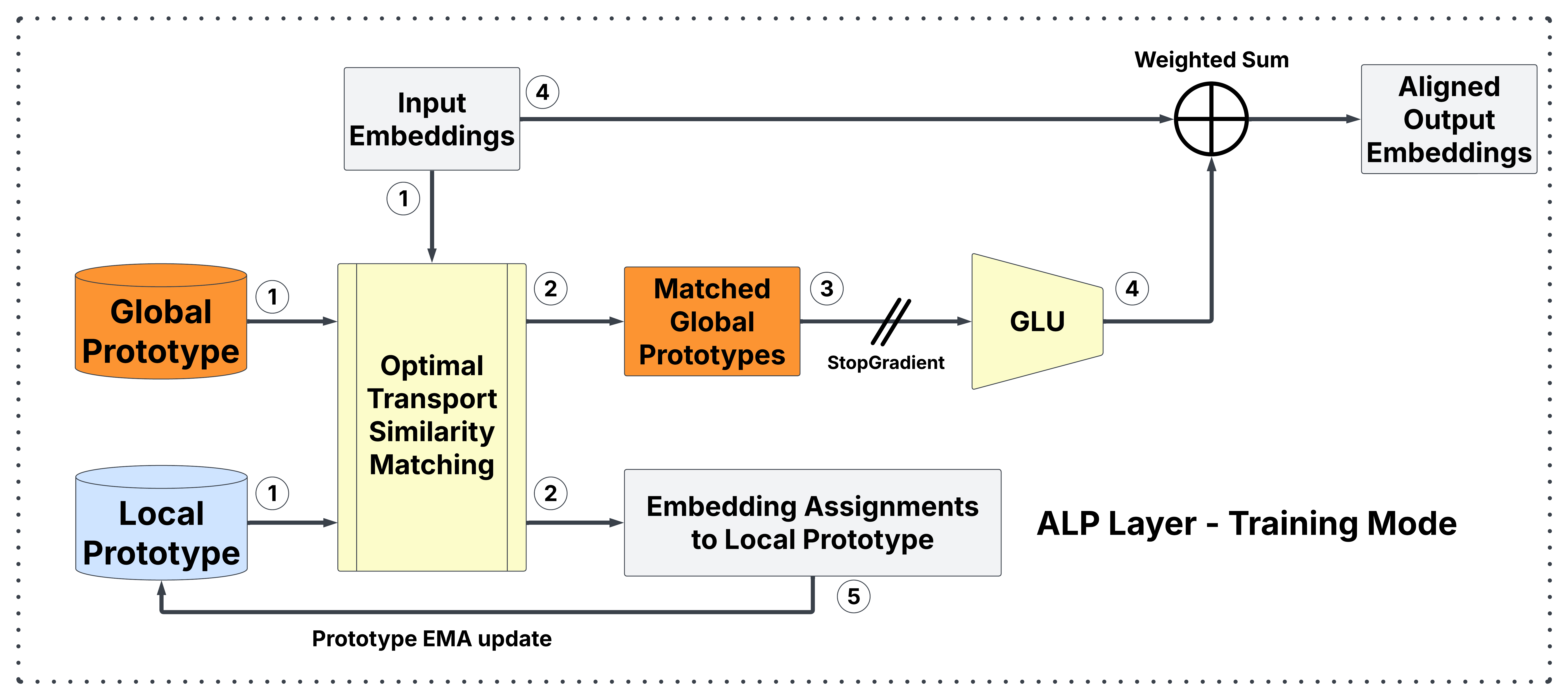}
\caption{The five processes of the ALP layer during training mode}
\label{Fig:alp_train}
\end{figure*}



\subsection{ALignment with Prototype Layer}



The core functionality of the ALP layer is to efficiently solve an optimal transport problem, aligning the distribution of input embeddings with the distribution of prototypes. Based on the computed transport plan, the update mechanism differs between training and inference. During training, local prototypes are updated by moving toward the assigned embeddings, while global prototypes serve to align the embeddings. During inference, only the trained local prototypes are used to align embeddings, adapting them to the client's local data distribution.

As shown in Figure \ref{Fig:ALP_position}, the ALP layer is placed at the beginning of the transformer encoder block, immediately before the Multi-Head Attention layer. Placing it here ensures that alignment benefits are introduced early, leading to more consistent feature embeddings throughout the encoder.

In transformer-based architectures, we assume that the ALP layer receives a sequence of embeddings \( \{x_b\}_{b=1}^B \), where each \( x_b \in \mathbb{R}^{Z,d} \) represents \( Z \) frames (for HAR) or patches (for vision), with an embedding dimension \( d \). To process all embeddings at once for efficiency gains, rather than sequentially for each batch, we reshape the input by merging the batch and sequence dimensions, transforming it from \( \mathbb{R}^{B, Z, d} \) to \( \mathbb{R}^{(B \cdot Z),d} \).

At this stage, the operations of the ALP layer diverge depending on whether the model is in training or inference mode. During training, the ALP layer follows a five-step process, as illustrated in Figure \ref{Fig:alp_train}, whereas in inference mode, the process consists of only four steps since the local prototype update is omitted. 

In Step 1, prototypes are matched to the embeddings using an optimal transport plan. Step 2 computes the embedding-to-prototype assignments based on the transport plan. Step 3 refines the selected prototypes before applying them to the embeddings. In Step 4, the embeddings are aligned with their assigned prototypes. Finally, Step 5, performed only during training, updates local prototypes based on their assigned embeddings. The complete process is described below:




\paragraph{1. Matching Prototypes.} The embedding matrix \( x \) is subsequently mapped to sets of local and global prototypes. Specifically, we define local prototypes \( P_{local} \in \mathbb{R}^{G, d} \) to capture personal embedding semantics and global prototypes \( P_{global} \in \mathbb{R}^{G, d} \) to capture shared general embedding semantics between different clients, where \( G \) represents the number of prototypes. 



\begin{algorithm}
\caption{Alignment with Prototype (ALP) Layer}
\label{algo:ALP}
\begin{algorithmic}[1] 
\State \textbf{Inputs:} Input embeddings $X$, Local prototypes $P_{\text{local}}$, Global prototypes $P_{\text{global}}$, Alignment coefficient $\beta$, Prototype update rate $\gamma$
\State \textbf{Output:} Aligned embeddings $\hat{X}$

\If{training} 
    \State $O_{\text{local}}, O_{\text{global}} \gets$ compute optimal transport plan 
    \Statex \hspace{1.5em} from $X$, $P_{\text{local}}$, and $P_{\text{global}}$ using \eqref{eq:matching}
    
    \State $\hat{P}_{\text{global}} \gets$ retrieve matched global prototypes 
    \Statex \hspace{1.5em} from $O_{\text{global}}$
    
    \State $P_{\text{local}} \gets$ update local prototypes using $O_{\text{local}}$ 
    \Statex \hspace{1.5em} and $X$ following \eqref{eq:localUpdate}
    \State $P_{\text{GLU}} \gets$ apply GLU gating to $\hat{P}_{\text{global}}$

\Else
    \State $O_{\text{local}} \gets$ compute optimal transport plan 
    \Statex \hspace{1.5em} from $X$ and $P_{\text{local}}$ using \eqref{eq:matching}
    
    \State $\hat{P}_{\text{local}} \gets$ retrieve matched local prototypes 
    \Statex \hspace{1.5em} from $O_{\text{local}}$
    \State $P_{\text{GLU}} \gets$ apply GLU gating to $\hat{P}_{\text{local}}$

\EndIf

\State $\hat{X} \gets$ compute aligned embeddings using $X$ and $P_{\text{GLU}}$ with \eqref{eq:align}
\State \Return L2-normalized $\hat{X}$

\end{algorithmic}
\end{algorithm}



Next, we compute the assignment of embeddings to prototypes by matching their distributions. Since our alignment process is embedded as a layer, it must remain flexible and operate without requiring class labels. To achieve this, we leverage the Sinkhorn-Knopp algorithm \citep{cuturi2013sinkhorn}, which approximates an optimal transport plan to map the two distributions.

Calculating the transport plan begins with the construction of an initial similarity matrix \(O_0\), derived from the dot product of L2-normalized input embeddings \(x_{\text{norm}}\) and L2-normalized prototypes \(P_{\text{norm}}\). This matrix is then iteratively refined into an optimal transport plan by applying row and column normalization steps, ensuring that the transport matrix satisfies marginal constraints.

\begin{equation}\label{eq:matching}
\begin{split}
O_{0} &= \exp \left( \dfrac{ x_{norm} \: P_{norm}^{\top} } {\epsilon}\right) \\
O_{i+1} &= \text{Diag}(u_{i}) \:  O_{i} \:  \text{Diag}(v_{i})
\end{split}
\end{equation}

Where $u$ and $v$ are scaling vectors used to normalize the rows and columns of the similarity matrix $O$ to ensure that the matrix is doubly stochastic. Each row in the transport plan \( O \) quantifies the similarity associated with transporting a feature embedding to sets of prototypes $P$.

During training, the transport plan \( O \in \mathbb{R}^{B \cdot Z, 2G} \) encodes the similarity in assigning input embedding to both local prototypes \( P_{local} \) and global prototypes \( P_{global} \). As two prototypes will be used for different purposes,  we partition \( O \) into \( O_{local} \) and \( O_{global} \).

When in inference mode, only the local prototype is matched with the embeddings to produce the transport plan $O_{local}$, improving computational efficiency.


\paragraph{2. Extracting Matched Prototypes.} From \( O_{global}\) during training,  we derive hard assignments by selecting the global prototype with the highest similarity (lowest transportation cost) for each input embedding, producing a list of global prototypes matched to each respective embeddings \(\hat{P}_{global}\). Meanwhile, $O_{local}$, on the other hand, will be used later in Step 5 to update the local prototypes.

During inference, the matched local prototypes \( \hat{P}_{\text{local}} \) are obtained from \( O_{\text{local}} \) using the same similarity hard assignment strategy.





\paragraph{3. Regulating Prototypes}  
The matched prototypes are then processed by a Gated Linear Unit (GLU) layer \citep{dauphin2017language}, allowing the model to regulate each prototype's influence on the final output, producing $P_{GLU}$. Since prototypes are not updated via gradients, we apply a stop-gradient function below the GLU layer during training. This ensures that the optimal transport plan computation does not affect backpropagation, preventing redundant gradient computation to the prototypes.


\paragraph{4. Influencing Embeddings with Prototypes} Finally, $P_{GLU}$ is applied to the original input projections to effectively align them by:
\begin{equation}\label{eq:align}
\hat{x} = \beta P_{GLU} + (1 - \beta) x
\end{equation}
Where \( \beta \) is the weighting factor controlling the influence of the prototypes on the input embeddings. Finally, the aligned embeddings $\hat{x}$ are reshaped to their original input dimensions, L2-normalized, and outputted from the layer.

\paragraph{5. Updating Prototypes}  

During local training, only the local prototypes are updated, while global prototypes are updated on the server using FedAli's aggregation strategy, which is later detailed in the following section.  

The goal of local prototype updates is to ensure that they accurately represent the unique characteristics of each client's local dataset. Since the number of prototypes may be smaller than the number of feature embeddings, each prototype must be assigned multiple embeddings while preserving alignment quality.  

To achieve this, we transpose the embedding-to-prototype similarity matrix \( O_{\text{local}} \) to obtain a prototype-centric similarity assignment \( O_{\text{local}}^\top \). This allows each prototype to quantify its similarity to all embeddings. For each prototype, we select $k$ embeddings with the highest similarity scores (Top-K). Here, the amount of embeddings assigned to each prototype ($k$) is calculated by \( \lceil (B \cdot Z) / G \rceil \) to achieve complete embedding coverage by local prototypes. Once selected, their similarity scores are normalized so that their sum equals 1. 

Using these normalized similarity scores as weights, we compute a weighted average of the assigned embeddings, denoted as \( \bar{x} \), which serves as the target update for each local prototype. The local prototype update follows an EMA strategy and is defined as:


\begin{equation}\label{eq:localUpdate}
P_{local}^{t+1} = \gamma P_{local}^t + (1 - \gamma) \bar{x}
\end{equation}  

In cases where only one embedding is assigned to a prototype (i.e., \( \lceil (B \cdot Z) / G \rceil = 1 \)), the update simplifies to an EMA step between the prototype and its assigned embedding, without computing a weighted average.

The complete process of the ALP layer is detailed in the pseudocode presented in Algorithm \ref{algo:ALP}.

\subsection{Federated Alignment (FedAli)}\label{fedDALI}

\begin{algorithm}
\caption{Federated Alignment (FedAli)}
\label{algo:FedAli}
\begin{algorithmic}[1] 
\State \textbf{Inputs:} $C$ client, $R$ communication rounds, $L$ encoder blocks
\State Initialize $w_{1}$, $P_{global\_1}$
\For{each communication round $r = 1, 2 \ldots, R$}
    \State Server sends $w_{r},P_{global\_r}$ to all clients
    \ForParallel{c \in C}
        \State $w_{r+1}^{c}, P_{local}^c \gets$ \text{localTrainALP($w_{r}$,$P_{global\_r}$)}
    \EndFor
    \State $w_{r+1} , P_{centroid}\gets \sum_{c=1}^{C}\dfrac{n_c}{n} w_{r+1}^{c}$
    \For{each encoder block $l = 1, 2, \ldots, L$}
        \State $P_{\text{local}}^l \gets \{ P_{\text{local}}^{l\_c} \mid c \in [1, C]\}$
        \vspace{0.1cm}
        \State  $P_{global\_{r+1}}^l \gets \text{KmeanFit}(P_{local}^l, P_{centroid}^{l}$)

    \EndFor
    \State $P_{\text{global}_{r+1}} \gets \{P_{\text{global}_{r+1}}^l \mid l \in [1, L]\}$
\EndFor
\end{algorithmic}
\end{algorithm}


Here, we detail FedAli to coordinate client prototypes and the aggregation process. As prototypes can be considered as part of the model weights $P \subseteq w$, FedAli aggregation can seamlessly extend from the FedAvg aggregation process, with the addition of prototype management techniques. Thus, we only detail the prototype aggregation pipeline, where after local training, each client sends their updated local weights $w^c_{r+1}$ and local prototypes $P^c_{local}$ to the central server.

As clients in heterogeneous environments tend to update their local prototypes differently to adapt to their data, coordinate-wise weighted averaging, as used in other studies \citep{tan2022fedproto,xu2023personalized}, can combine conflicting local prototypes, leading to less generalized global prototypes. To address this, we cluster the aggregated local prototypes using K-mean \citep{lloyd1982least}, with the number of clusters set to \(G\). Furthermore, to improve the stability of the global prototypes throughout the next round, the initial K-mean centroid is computed through the results of the weighted average of the client's local prototypes $P_{centroid}$. This process is done for all $L$ ALP layers within the client's model architecture.


The convergent centroids, which are now the global prototypes for the next communication round \( P_{global_{r+1}} \), along with the new global model $w_{r+1}$ obtained from the weighted average of the client model, are sent back to the clients. Note that, to gain communication cost, local and global prototypes are never communicated together (only the updated local prototypes are aggregated, and the new round global prototypes are sent down). The complete aggregation process in FedAli is detailed in Algorithm \ref{fedDALI}.

\section{Evaluations}

The experiments, as has been done similarly in previous studies \citep{10.1145/3570361.3592505,cai2023federated}, were emulated on a high-performance computing cluster. We utilize 8 Nvidia's V100 GPUs, each simulating multiple local clients to parallelize the distributed training process. The HAR experiments were implemented using TensorFlow, whereas the vision experiments were performed using PyTorch. For reproducibility and further research, we publicly release the source code of our work\footnote{https://github.com/getalp/FedAli}.

\subsection{Experimental Setting} \label{sec:setup}

We train each FL strategy for 600 communication rounds for HAR tasks and 200 communication rounds for vision tasks. Both tasks are trained using an Adam optimizer with a learning rate of $1 \times 10^{-4}$, where each client is trained for $5$ local epochs with a batch size of $64$. 

For all clients in our study, we use a class-stratified partitioning strategy to preserve the inherent class imbalance across both the training and the testing datasets. Specifically, each local dataset is split into 80\% training data and 20\% testing data.

\subsubsection{HAR Tasks}

To simulate diverse and realistic real-world heterogeneity (non-IID data) among FL clients, we compare and evaluate FedAli using the following HAR datasets, each reflecting different forms of data heterogeneity influenced by various factors:

\begin{itemize}
\item \textbf{HHAR} dataset \citep{stisen2015smart} represents a data heterogeneous learning environment due to the variety of sensing devices. Each of the 9 subjects wore 8 Android smartphones (2 LG Nexus 4, 2 Samsung Galaxy S3, 2 Samsung Galaxy S3 mini, and 2 Samsung Galaxy S plus) in a tight pouch carried around the waist while performing 6 different activities (biking, sitting, standing, walking, upstairs and downstairs). The recordings of the subjects and multiple smartphones totaled around 26 hours.

\item \textbf{RealWorld} dataset \citep{realword} shows a data heterogeneous learning environment caused by differences in the location of the on-body wearable sensor. Each of the 15 diverse subjects is mounted with six Samsung Galaxy S4 smartphones and an LG G Watch R placed in 7 different body positions: head, chest, upper arm, waist, forearm, thigh, and shin. The subjects performed various outdoor activities without restrictions and the data was labeled into eight activities. Downstairs, Upstairs, Lying, Sitting, Standing, Walking, Jumping, and Running. From the different devices and subjects, there is an approximate total of 130 hours of recorded data.

\item \textbf{Combined} is made up of the HHAR and RealWorld to simulate heterogeneity over several factors between multiple FL clients with different activities and amount of training data. Specifically, this consists of differences in subjects, environments, devices, positions, activities, and their class imbalance between each client.
\end{itemize}

To generate clients with the HAR datasets, we do not rely on artificial client partitioning. Instead, we treat each device from a unique subject as an independent client in our distributed experiments. This setup resulted in 36 clients for HHAR (where data from two identical smartphone models were considered as one), 105 clients for RealWorld, and 141 clients for the combined dataset setup. For the classification tasks, we used the lightweight transformer HART \citep{ek2023transformer}, which consists of six encoder blocks and a projection size of 192.

Since the devices in the two datasets were recorded at different sampling rates, we first applied an anti-aliasing filter, downsampling all signals to 50 Hz. Next, we retained only the raw signals, which were normalized using sensor-wise z-normalization for each device. We then segmented the data into sliding windows of 128 samples (2.56 s) with 50\% overlap. Each window encompassed six sensor channels: three from the accelerometer and three from the gyroscope.

\subsubsection{Vision Tasks}
We evaluate our method using the community vision benchmark datasets below: 

\begin{itemize} \item \textbf{CIFAR-10} dataset \citep{krizhevsky2009learning} consists of 60,000 color images of size 32×32 pixels, evenly distributed across 10 object classes: airplane, automobile, bird, cat, deer, dog, frog, horse, ship, and truck. Each class contains 6,000 images, with 50,000 images used for training and 10,000 for testing.
\item \textbf{CIFAR-100} dataset \citep{krizhevsky2009learning} extends CIFAR-10 with a more fine-grained classification task. It contains 100 classes grouped into 20 superclasses, where each class consists of 500 training images and 100 testing images. The dataset retains the same structure as CIFAR-10, but introduces greater complexity and intraclass variation.
\end{itemize}

To simulate a non-IID data distribution among FL clients, we applied a Dirichlet-based partitioning strategy \citep{marfoq2022personalized, li2023fedtp}, setting the partitioning sharpness parameter to 0.3, while generating 20 and 100 clients for each of the two vision datasets, respectively. We employed a tiny ViT with six encoder blocks and a projection size of 192 as the backbone model.

\begin{table*}[ht!]
    \centering
    \caption{Comparison of FL Strategies on HHAR, RealWorld, and Combined Datasets trained over 600 CR. Results in \textbf{bold} represent the best performance, while results \uline{underlined} represent the second-best performance.}
    \resizebox{\linewidth}{!}{
    \begin{tabular}{lccc|ccc|ccc}
        \toprule
        \multirow{2}{*}{} & \multicolumn{3}{c|}{\textbf{HHAR (36 Clients)}} & \multicolumn{3}{c|}{\textbf{RealWorld (105 Clients)}} & \multicolumn{3}{c}{\textbf{Combined (141 Clients)}} \\
        & \textbf{Personalization} & \textbf{Generalization} & \textbf{Global} & \textbf{Personalization} & \textbf{Generalization} & \textbf{Global} & \textbf{Personalization} & \textbf{Generalization} & \textbf{Global} \\
        \midrule
        Centralized & N/A & N/A & 97.55 & N/A & N/A & 89.78 & N/A & N/A & 91.36 \\
        Local & 93.84 ± 7.14 & 33.75 ± 9.21 & N/A & 91.44 ± 6.06 & 18.67 ± 3.58 & N/A & 91.84 ± 6.51 & 14.20 ± 3.05 & N/A \\
        \cmidrule{1-10}
        FedAvg & 96.88 ± 3.46 & 75.85 ± 4.97 & 86.47 & 90.72 ± 5.85 & 60.41 ± 3.85 & 72.47 & 91.46 ± 6.59 & 56.24 ± 4.29 & 68.78 \\
        FedPer & 95.89 ± 4.68 & 52.32 ± 9.99 & N/A & 89.52 ± 7.21 & 23.10 ± 3.71 & N/A & 88.12 ± 7.76 & 17.41 ± 3.36 & N/A \\
        FedProx & 97.35 ± 2.67 & 74.97 ± 5.02 & 86.87 & 90.97 ± 5.67 & \uline{60.92} ± 3.69 & \uline{73.45} & 91.52 ± 6.65 & \uline{55.84 ± 4.22} & 68.56 \\
        MOON & 97.16 ± 3.63 & \uline{75.44 ± 5.47} & \uline{88.03} & 91.03 ± 5.74 & 60.26 ± 3.98 & \textbf{74.30} & \uline{91.86 ± 6.26} & 55.04 ± 4.85 & \textbf{70.00} \\
        FedProto & 92.51 ± 9.92 & 41.66 ± 10.29 & N/A & \textbf{91.55} ± 6.26 & 21.54 ± 3.69 & N/A & 91.63 ± 7.59 & 16.00 ± 3.36 & N/A \\
        FedPAC & \uline{97.58 ± 2.58} & 73.05 ± 6.95 & N/A & \uline{91.06 ± 6.06} & 34.94 ± 5.99 & N/A & \textbf{91.91 ± 5.97} & 24.55 ± 5.53  & N/A \\
        FedAli & \textbf{98.15 ± 1.75} & \textbf{81.84 ± 4.81} & \textbf{90.68} & 90.333 ± 5.77 & \textbf{61.33 ± 3.66} & 71.50 & 90.80 ± 6.67 & \textbf{57.25 ± 4.29} & \uline{68.94} \\
        \bottomrule
    \end{tabular}
    }
    \label{tab:results}
\end{table*}

\subsubsection{Baselines} 
We compare FedAli against two conventional learning approaches and six federated learning strategies. The baselines are detailed as follows:
\begin{itemize}
\item \textbf{Centralized} approach assumes that the server has access to all client data and trains the model for 200 epochs. 
\item \textbf{Local} refers to the setup in which clients train independently on their local dataset for 200 epochs without any communication with others.
\item \textbf{FedAvg} \citep{mcmahan2017communication} aggregates local model updates using a weighted average, serving as the foundation for most FL methods.
\item \textbf{FedPer} \citep{arivazhagan2019federated} introduces personalized local layers, where the lower layers of the model are shared between clients while the upper layers remain client-specific to adapt to local data distributions.
\item \textbf{FedProx} \citep{li2020federated} extends FedAvg by adding a proximal term that constrains diverging local updates. We tuned this term to values of $0.01$, $0.1$, and $1.0$, finding $0.01$ to gave the best overall performance.
\item \textbf{MOON} \citep{li2021model} leverages contrastive learning to align local models with the global model. We tuned the contrastive loss coefficient to $0.01$, $0.1$, and $1.0$, with $1.0$ performing the best.
\item \textbf{FedProto} \citep{tan2022fedproto} employs class prototype-based representation learning, where clients share class prototypes instead of full model updates to enforce similarity. Similarly, we tuned the contrastive loss coefficient and found 1.0 to be optimal.

\item \textbf{FedPAC} \citep{xu2023personalized} is another prototype-based approach that aligns feature representations between clients for consistency while introducing a novel method for personalized classifier collaboration.
\end{itemize}



For FedAli, after tuning (Section \ref{sec:sensitivity}), we set the alignment coefficient $\beta = 0.2$ and the prototype update rate $\gamma = 0.999$. Given that the lower layers capture various local features while the upper layers focus on more abstract global representations \cite{shin2024effective}, we structure the number of prototypes to decrease across the six encoder blocks, starting from $G = {2048, 1024, 512, 256, 128, 64}$.  For the Sinkhorn-Knopp algorithm, following \cite{caron2020unsupervised}, we set the sharpness parameter to $\epsilon = 0.05$ and use three iterations of the normalization step to balance between performance and efficiency.

\subsubsection{Metrics}

To evaluate the performance of FedAli and the baseline methods, we utilized the following metrics:

\begin{itemize}
\item \textbf{Personalization score} represents the averaged macro F1 scores of all clients on their respective local test sets. The client models used for this evaluation are selected from the communication round that achieved the highest average accuracy.
\item \textbf{Generalization score} assesses the average performance of each client model on the test sets of all participating clients. The same models used for the Personalization Score evaluation are employed here.
\item \textbf{Global score} evaluates the global model on the test sets of all clients. For this assessment, we used the global model from the communication round that achieved the highest accuracy.
\end{itemize}

\subsection{Results on HAR and Vision}\label{sec:Results}

The performance of multiple FL strategies on the HHAR, RealWorld, and the combination of the two is reported in Table \ref{tab:results}. Regarding the overall compared FL approaches, the results consistently show that FedAli excels in generalization with strong personalization and global performance across different datasets, highlighting its robustness in FL environments, especially on the HHAR dataset, where FedAli emerges as the top-performing strategy with a large margin over other approaches, showing the best adaptability across all metrics. MOON performs well across all datasets, particularly excelling in the RealWorld and Combined datasets for global performance. FedAvg and FedProx consistently deliver balanced results, while FedProto and FedPer show limitations in generalization, often leading to overfitting. FedPAC, while competitive in terms of personalization across all datasets, declined in terms of generalization. The Local strategy, despite strong personalization, consistently underperforms in generalization, highlighting the trade-off between local adaptability and broader applicability. Although the centralized results can exhibit the best overall results (acts as the upper bound) in our findings here, they are not directly comparable to the FL strategies due to the need to communicate and aggregate user data. 

\begin{table*}[ht!]
    \centering
    \caption{Comparison of FL Strategies on CIFAR10 and CIFAR100 with 20 and 50 clients trained with 200 CR. Results in \textbf{bold} represent the best performance, while results \uline{underlined} represent the second-best performance.}
    \resizebox{\linewidth}{!}{
    \begin{tabular}{l|ccc|ccc|ccc|ccc}
        \toprule
        \multirow{2}{*}{} & \multicolumn{3}{c|}{\textbf{CIFAR10 (20 Clients)}} & \multicolumn{3}{c|}{\textbf{CIFAR10 (50 Clients)}} & \multicolumn{3}{c|}{\textbf{CIFAR100 (20 Clients)}} & \multicolumn{3}{c}{\textbf{CIFAR100 (50 Clients)}} \\
        & \textbf{Per.} & \textbf{Gen.} & \textbf{Global} & \textbf{Per.} & \textbf{Gen.} & \textbf{Global} & \textbf{Per.} & \textbf{Gen.} & \textbf{Global} & \textbf{Per.} & \textbf{Gen.} & \textbf{Global} \\
        \midrule
        FedAvg & \uline{76.91 ± 6} & \textbf{43.29 ± 7} & \uline{64.41} & 78.29 ± 11 & 34.62 ± 9 & \textbf{59.78} & 42.88 ± 4 & 19.11 ± 1 & \uline{32.55} & 40.93 ± 5 & 17.39 ± 1 & 31.29 \\
        FedPer & 76.05 ± 6 & 39.57 ± 6 & N/A & 75.76 ± 12 & 29.64 ± 7 & N/A & 37.19 ± 5 & 11.39 ± 1 & N/A & 30.336 ± 4 & 8.34 ± 1 & N/A \\
        FedProx & 75.14 ± 5 & \uline{43.21} ± 6 & 61.94 & \uline{78.57 ± 10} & \textbf{35.64 ± 7} & \uline{57.88} & 41.66 ± 4 & \uline{20.47 ± 1} & 30.90 & \uline{42.04 ± 4} & \textbf{18.84 ± 1} & \uline{31.49} \\
        MOON & 63.08 ± 8 & 25.14 ± 3 & 18.44 & 66.79 ± 15 & 19.44 ± 5 & 20.52 & 29.40 ± 5 & 7.48 ± 1 & 4.48 & 25.08 ± 4 & 
        5.45 ± 1 & 4.22 \\
        FedProto & 69.16 ± 6 & 29.60 ± 4 & N/A & 70.61 ± 14 & 22.61 ± 5 & N/A & 30.99 ± 6 & 9.24 ± 1 & N/A & 22.21 ± 5 & 5.99 ± 1 & N/A \\
        FedPAC & 74.66 ± 6 & 38.45 ± 6 & N/A & 76.80 ± 11 & 30.87 ± 8 & N/A & \uline{45.33 ± 4} & 16.50 ± 1 & N/A & 38.66 ± 4 & 16.23 ± 1 & N/A \\
        FedAli & \textbf{77.68 ± 5} & 43.14 ± 6 & \textbf{64.46} & \textbf{79.89 ± 10} & \uline{34.90 ± 9} & 57.39 & \textbf{46.49 ± 5} & \textbf{21.24 ± 1} & \textbf{35.45} & \textbf{43.70 ± 4} & \uline{18.30 ± 1} & \textbf{32.71} \\
        \bottomrule
    \end{tabular}
    }
    \label{tab:vision_results}
\end{table*}

 The comparative results in Table \ref{tab:vision_results} show that FedAli, also within vision tasks, outperforms most previous PFL baselines. FedProx exhibits strong generalization, particularly with a larger number of clients across both datasets. FedPAC also shows competitive performance, especially in the personalized evaluation of CIFAR100. In contrast, FedPer and FedProto yield limited results in heterogeneous settings, particularly in generalization. Although FedAvg performs well on CIFAR10, it struggles on CIFAR100 due to increased data heterogeneity. In general, FedAli offers a strong balance between generalization and personalization, highlighting its effectiveness in vision tasks.

 \subsection{Qualitative analysis with t-SNE} 

We present t-SNE visualizations \citep{van2008visualizing} to illustrate how alignment of our proposed prototypes enhances the robustness of local models during inference and reduces divergence between clients during training. First, we highlight the misalignment issue in standard FL methods using FedAvg, followed by a comparison with the embeddings produced by FedAli. We conducted two experiments:

\textbf{Inference: Local vs. Unseen Embeddings} In this setup, we select a HHAR client from our Combined setup and evaluate the local model on both its own local test-set and a RealWorld client test-set that was never seen locally. These datasets share five common activities: Sitting, Standing, Walking, Upstair, and Downstair.

\begin{figure}[ht]
    \centering
    \begin{subfigure}[t]{0.49\linewidth}
        \centering
        \includegraphics[width=\linewidth]{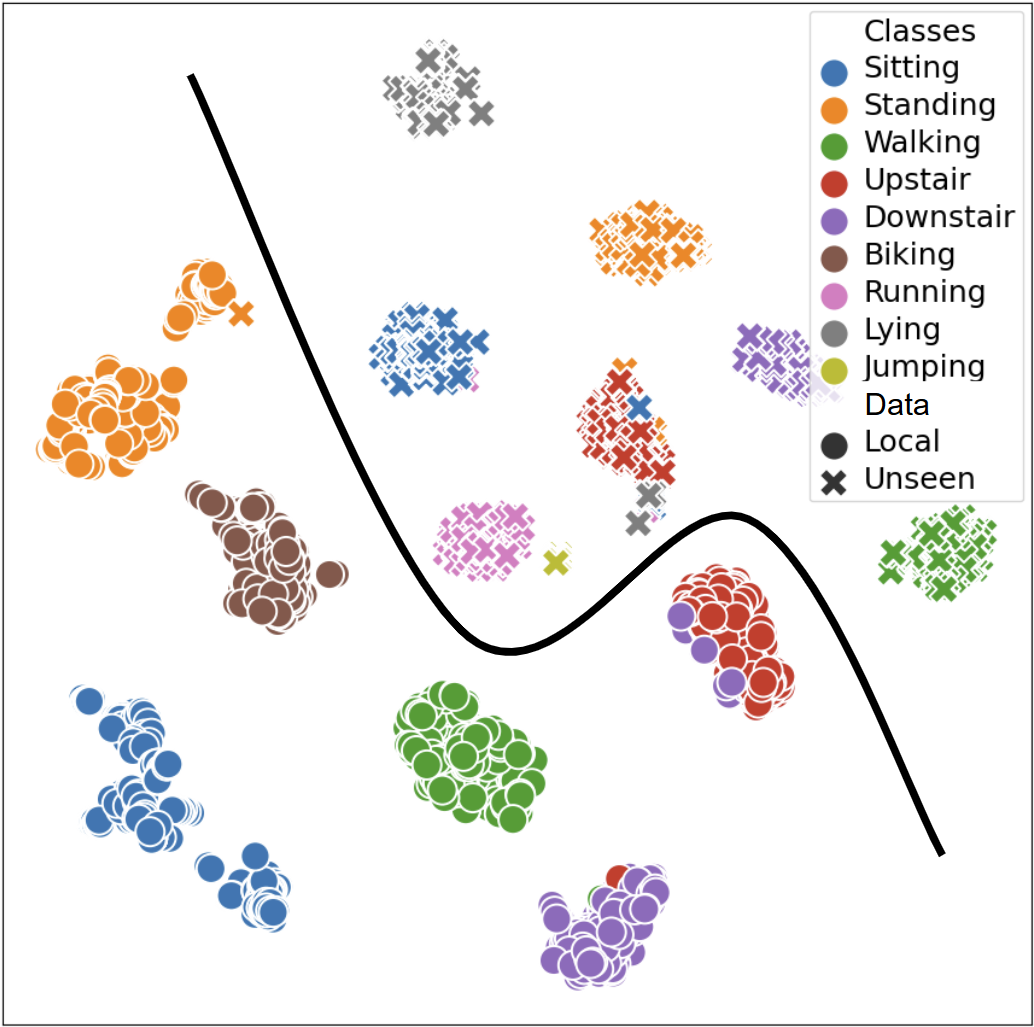}
        \caption{FedAvg: Unseen class representations does not share the same space as local ones.}
        \label{Fig:fedavg_inference}
    \end{subfigure}
    \hfill
    \begin{subfigure}[t]{0.49\linewidth}
        \centering
        \includegraphics[width=\linewidth]{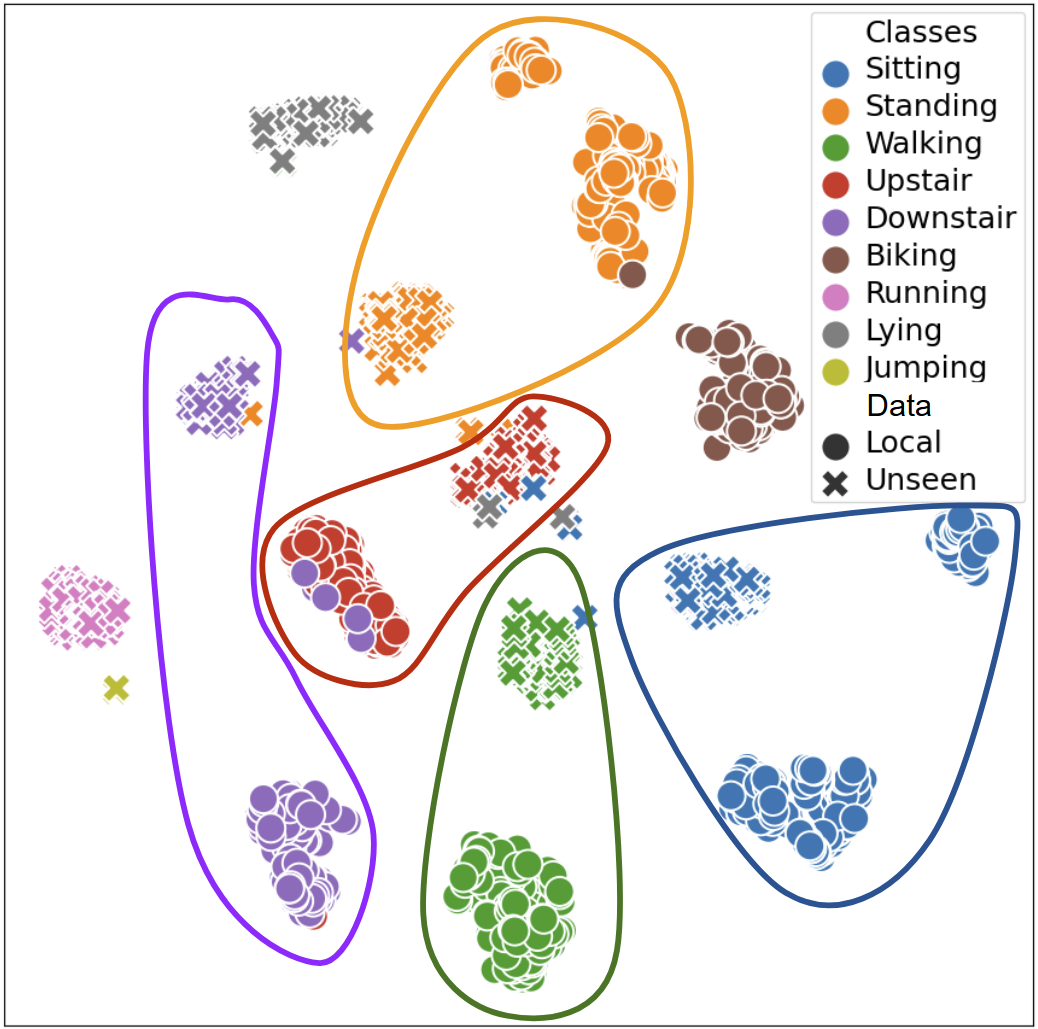}
        \caption{FedAli: Unseen class representations align more closely with local ones.}
        \label{Fig:fedali_infernece}
    \end{subfigure}
    
    \caption{t-SNE visualization comparing FedAvg (left) and FedAli (right). The inference representation of a HHAR client model on it's local data and the data of an unseen RealWorld client.}
    \label{Fig:FL_inference}
\end{figure}

Figure \ref{Fig:fedavg_inference} illustrates the embeddings produced by FedAvg, where there is a distinct separation between the activities of the two datasets. In contrast, Figure \ref{Fig:fedali_infernece} demonstrates that FedAli leads to better intermixing of embeddings of the same activities. In particular, FedAli creates observable unified class-level boundaries between the two different datasets, effectively transforming unseen embeddings from another environment to align more closely with the representation of the local model.



\begin{figure}[ht]
    \centering
    \begin{subfigure}[t]{0.49\linewidth}
        \centering
        \includegraphics[width=\linewidth]{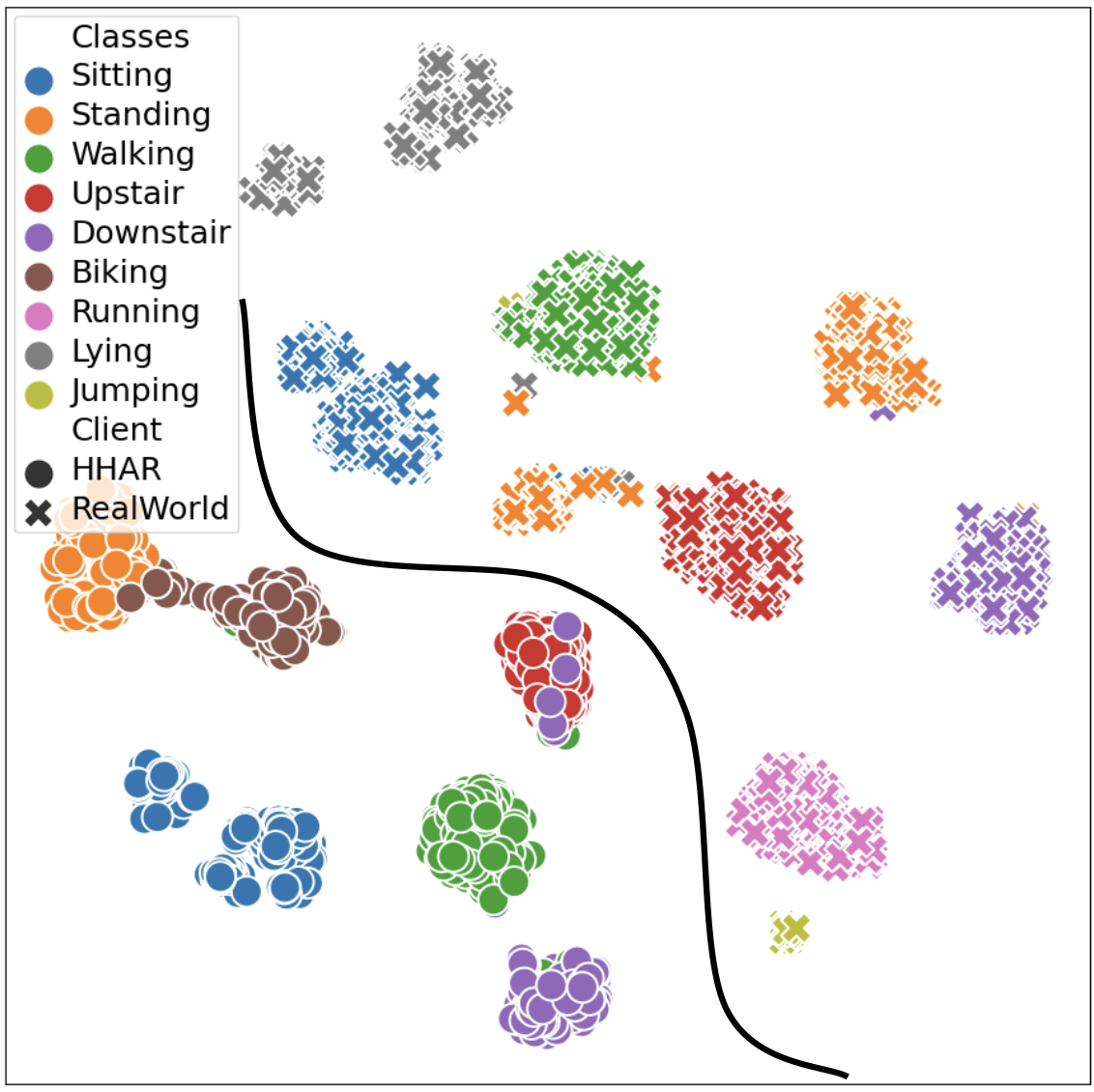}
        \caption{FedAvg: HHAR and RealWorld representations are separated by a clear boundary.}
        \label{Fig:fedavg_train}
    \end{subfigure}
    \hfill
    \begin{subfigure}[t]{0.49\linewidth}
        \centering
        \includegraphics[width=\linewidth]{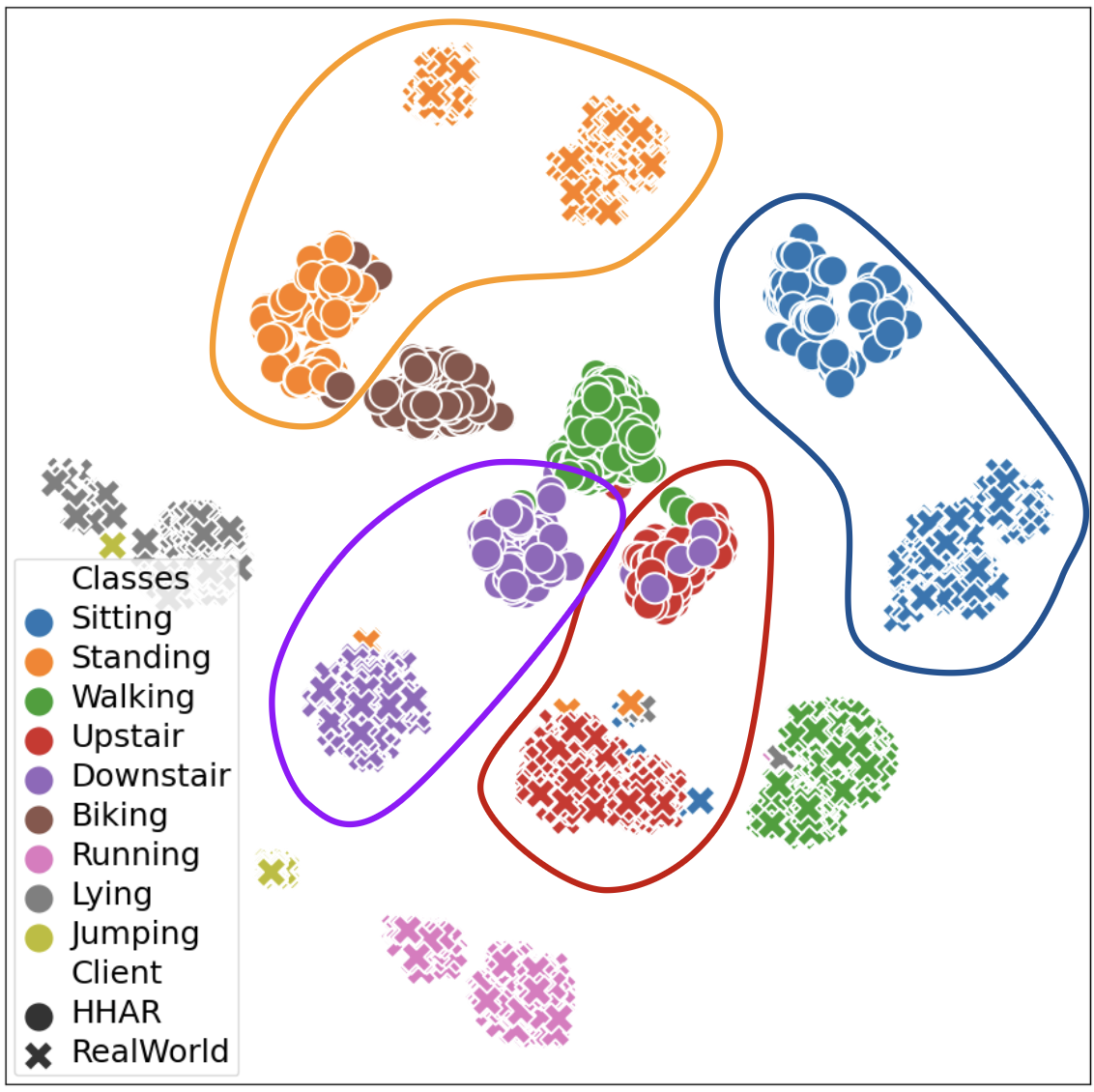}
        \caption{FedAli: 4 out of the 5 overlapping activities share a class boundary within their learned representation space.} 
        \label{Fig:fedali_train}
    \end{subfigure}
    
    \caption{t-SNE visualization comparing FedAvg (left) and FedAli (right). The training data representations of an HHAR client on its own dataset and a RealWorld client on its respective dataset, both plotted into the same representation space to show training alignment.}
    \label{Fig:FL_train}
\end{figure}

\textbf{Training: Embedding Alignment Across Clients} In this experiment, we compare the embeddings of two clients. One from HHAR and one from RealWorld from our Combined learning setup. Each client performs inference on its own local dataset, and we plot the resulting representations in a shared space. Ideally, in a well-aligned model within a distributed environment, the same classes should be mapped to similar regions in the embedding space.

Figure \ref{Fig:fedavg_train} shows the embeddings of FedAvg-trained clients, where a clear boundary here also separates the representations of the two datasets. In contrast, Figure \ref{Fig:fedali_train} reveals that FedAli improves alignment. Four out of five overlapping activities share a class boundary, indicating greater consistency in how activities are represented. However, the Walking activity remains misaligned, as it is split by Upstair embeddings. This discrepancy is reasonable, since the movements involved in Walking and Upstair can be quite similar. Additionally, since FedAli does not use explicit class labels for alignment, some limitations arise when two similar activities exhibit subtle but distinct similarities.

\label{para:pretrain}
 \subsection{Pre-trained initialization and convergences} 

 Here we show that, unlike other FL regularization strategies that employ class prototypes \citep{tan2022federated,tan2022fedproto,xu2023personalized}, the prototypes in FedAli are feature-based and do not rely on labels or the need to induce a new loss term during training. By incorporating the regularization through a layer embedded with the network, the model becomes flexible towards any tasks and the prototypes can be pre-trained in an unsupervised way.

The versatility of our proposal is imperative, as starting FL training from a pre-trained model has been a reliable approach to mitigate data heterogeneity and significantly increase the convergence rate \citep{Li2020On,nguyen2023where}. Specifically, pre-trained personalized client models make smaller learning steps during local fine-tuning, and their distance from the global server model is lessened. Furthermore, the relative abundance of unlabeled data compared to labeled data will often mean that the pre-training method will be performed using an unsupervised method \citep{plotz2023if}.



We pre-trained HART along with local prototypes in a self-supervised manner using MAE \citep{he2022masked} on five publicly available HAR datasets \cite{8418369,Malekzadeh:2018:PSD:3195258.3195260,reiss2012introducing,Anguita2013APD,vavoulas2016mobiact}. Afterwards, we extracted the pre-trained encoder and added a 1024-unit dense layer followed by the classification heads. The model was then fine-tuned using FedAli for 200 communication rounds. Before sharing the initialized classifier with clients, we set the pre-trained local prototypes as global prototypes. 

\begin{figure}[ht]
\centering
\includegraphics[width=0.65\linewidth]{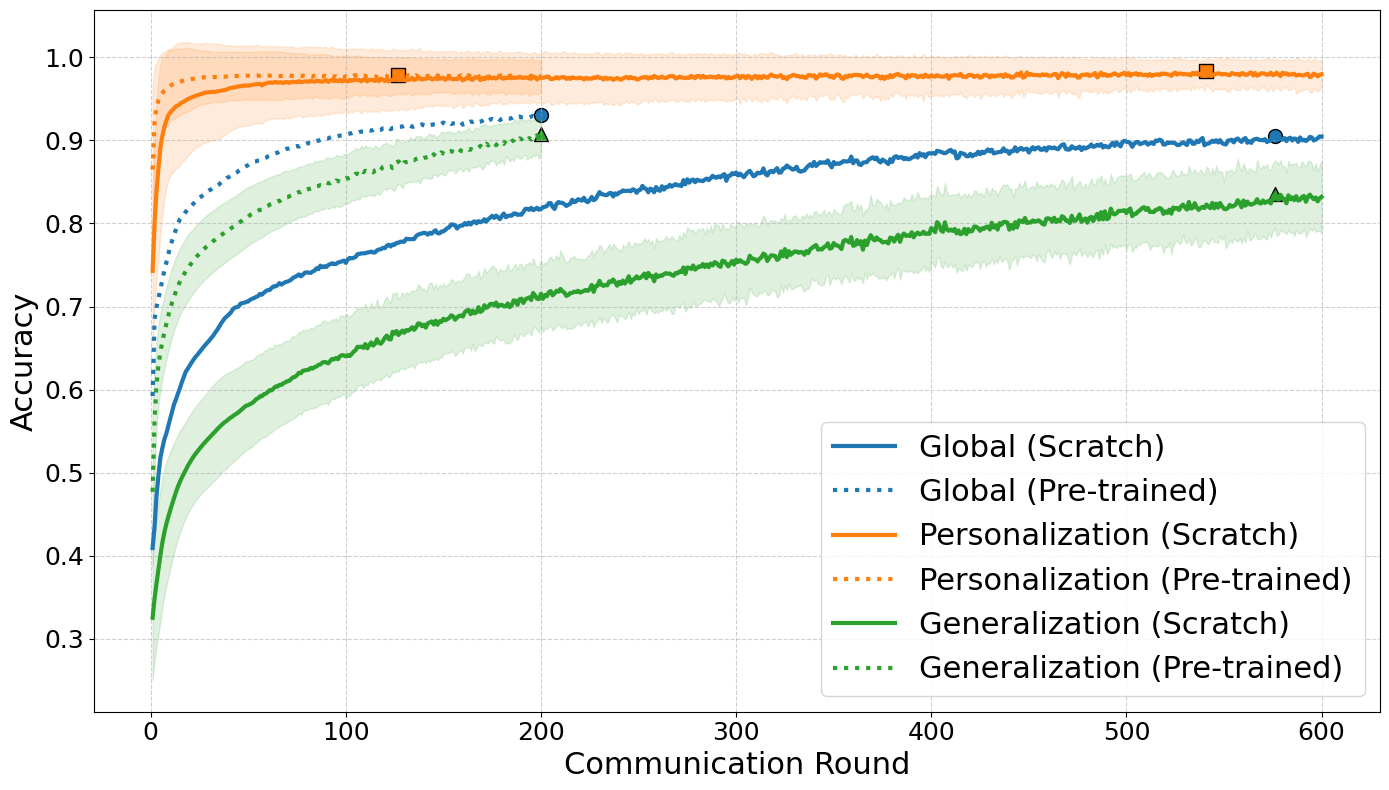}
\caption{Learning curve of comparison of starting from randomly initialized weights (Scratch) versus starting from a pre-trained model on the HHAR dataset. The marked shapes illustrate the highest point.}
\label{Fig:pretrain_learningcurve}
\end{figure}

First, we present the gains from starting from a pre-trained compared to learning from scratch. The results presented in Figure \ref{Fig:pretrain_learningcurve} show that FedAli, on HHAR clients, starting the FL from our pre-trained model with prototypes can significantly improve generalization (more than 20\% gain) and global performance (more than 11\% gain) while retraining personalization performance and reaching a smaller standard deviation. Additionally, the models achieve a much faster convergence rate, whereas training from scratch requires nearly 600 communication rounds for the model to converge.  

\begin{figure}[ht]
    \centering
    \begin{subfigure}[b]{0.3\textwidth}
        \centering
        \includegraphics[width=\textwidth]{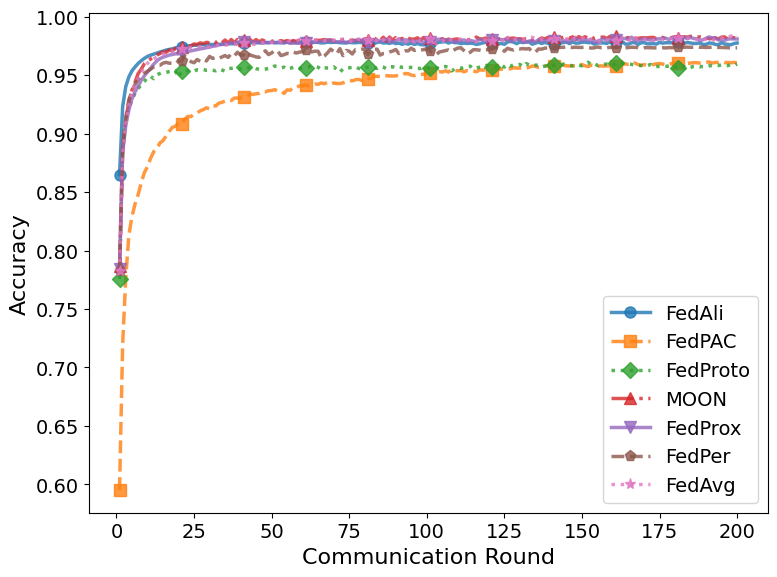}
        \caption{HHAR Personalization}
        \label{fig:sub1}
    \end{subfigure}
    \hfill
    \begin{subfigure}[b]{0.3\textwidth}
        \centering
        \includegraphics[width=\textwidth]{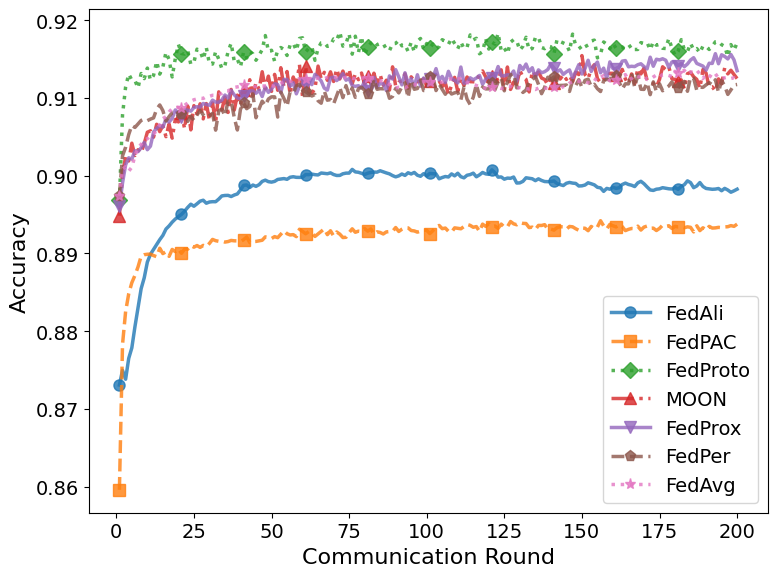}
        \caption{RealWorld Personalization}
        \label{fig:sub2}
    \end{subfigure}
    \hfill
    \begin{subfigure}[b]{0.3\textwidth}
        \centering
        \includegraphics[width=\textwidth]{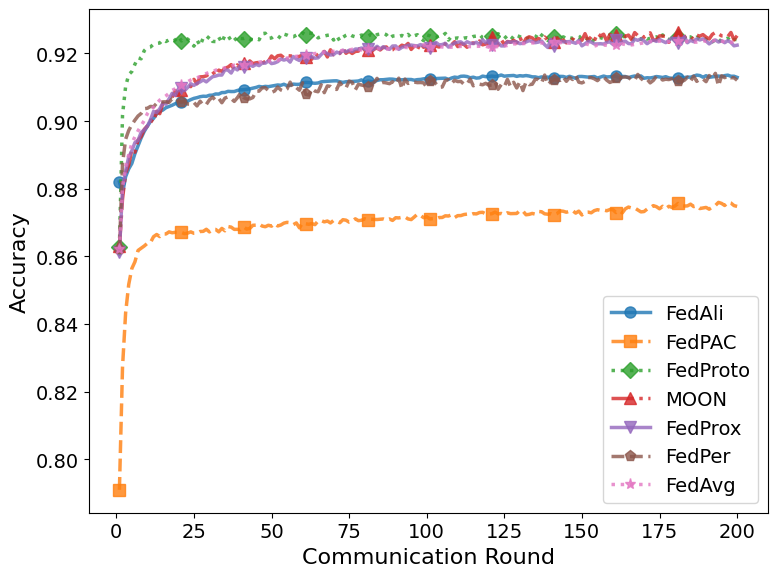}
        \caption{Combined Personalization}
        \label{fig:sub3}
    \end{subfigure}
    
    \vspace{1em} 

    \begin{subfigure}[b]{0.3\textwidth}
        \centering
        \includegraphics[width=\textwidth]{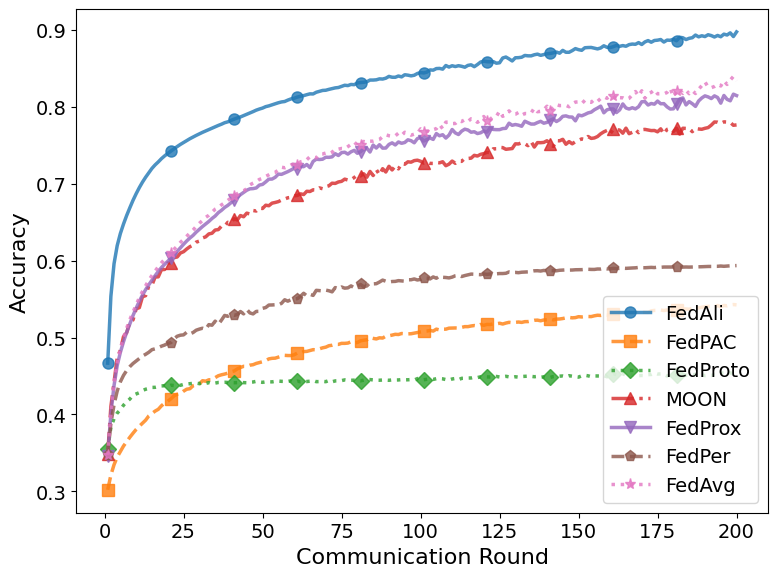}
        \caption{HHAR Generalization}
        \label{fig:sub4}
    \end{subfigure}
    \hfill
    \begin{subfigure}[b]{0.3\textwidth}
        \centering
        \includegraphics[width=\textwidth]{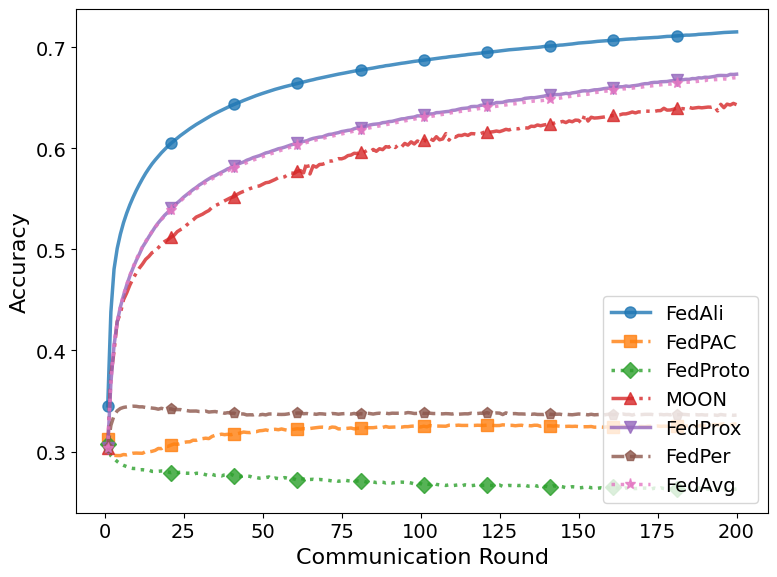}
        \caption{RealWorld Generalization}
        \label{fig:sub5}
    \end{subfigure}
    \hfill
    \begin{subfigure}[b]{0.3\textwidth}
        \centering
        \includegraphics[width=\textwidth]{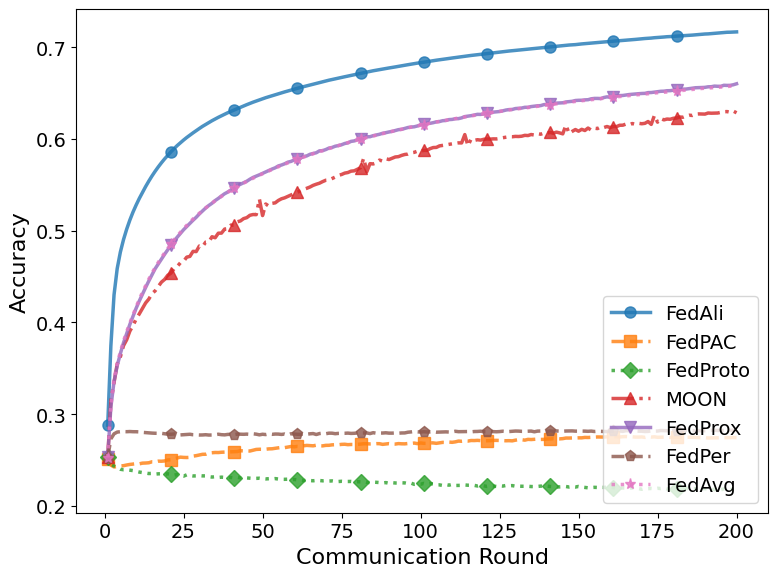}
        \caption{Combined Generalization}
        \label{fig:sub6}
    \end{subfigure}
    
    \caption{Learning curves of FL strategies on the HHAR, RealWorld and the Combined datasets.}
    \label{fig:comparison_curve}
\end{figure}

Next, we assess the impact of pre-training for all the other different FL strategies, and we repeated the pre-training process without the ALP layer. In Figure \ref{fig:comparison_curve}, we present the personalization and generalization learning curves for HAR datasets, comparing various FL strategies initialized from MAE pre-training. In terms of personalization, all approaches converge rapidly, with relatively small differences (within a single digit) in performance. However, when evaluating generalization, we observe a slower convergence rate and larger performance discrepancies (more than 10\%).

In particular, FedAli shows significantly superior generalization performance and convergence speed while maintaining competitive personalization performance compared to other FL strategies. In contrast, other prototype-based approaches, such as FedProto and FedPAC perform, are severely lacking in terms of client model generalization.

As discussed previously, relying on a single global prototype per class struggles to represent heterogeneous data effectively. This limitation is particularly evident in more challenging datasets, such as RealWorld (which varies in users and on-body sensor placements) and Combined (which introduces variations in users, devices, environments, and on-body locations). Furthermore, methods like FedPer and FedProto, which do not communicate the entire model within the FL framework, reduce the potential for knowledge transfer across client environments. Lastly, FedPAC, which prioritizes training the feature extractor over the classification heads, does not fully take advantage of the pre-trained feature extractor benefits, ultimately limiting its effectiveness.

\subsection{Communication and computation overhead}

Table \ref{tab:comparison} presents the training time, storage / memory requirement, and communication cost of all FL strategies, relatively, compared to FedAvg in our configurations. We show that while FedAli indeed has more parameters due to the prototypes, our proposal is still relatively much faster than other recent FL strategies that rely on contrastive techniques during training. Additionally, unlike other prototype-based methods, FedAli does not require clients to store all data representations (Data Rep.) from their local datasets to compute local prototypes, potentially reducing local storage and memory costs, especially at scale on clients with large amounts of data.

\section{Ablation}

The local and global prototypes can be mutually exclusive in our proposed alignment framework. To evaluate the effectiveness of the ALP layer, we performed an ablation study by removing specific components and assessing the impact on the performance of the HHAR clients. We conducted three random runs over 200 communication rounds for each study to ensure the stability of our findings.

The results in Table \ref{tab:ablation} confirm the necessity of each component in our framework, highlighting the crucial role of GLU and dual prototypes in the balance of personalization and generalization performance. Specifically, we observe that removing the global prototype's regularization increases the Personalization score but leads to a decline in both Generalization and Global scores. Likewise, when local prototypes are not utilized during inference, generalization performance deteriorates. 

\begin{table}[H]
    \vspace{-10pt} 
    \centering
    \caption{Local training time, memory footprint, and communication overhead relative to FedAvg with HHAR clients} 
    \resizebox{\linewidth}{!}{
    \begin{tabular}{lrrr}
        \toprule
        \textbf{Method} & \textbf{Local train time} $\downarrow$ & \textbf{ Memory footprint} $\downarrow$ & \textbf{Comm overhead} $\downarrow$ \\
        \midrule
        FedAvg    & 0.57 Minutes & 5.8 MegaBytes & 5.8 MegaBytes \\
        FedPer    & 1.00$\times$ & 1.00$\times$ & 0.14$\times$ \\
        FedProx   & 2.51$\times$ & 1.86$\times$ & 1.00$\times$ \\
        MOON      & 2.30$\times$ & 2.72$\times$ & 1.00$\times$ \\
        FedProto  & 2.28$\times$ & Data Rep. + 1.00$\times$ &   0.01$\times$ \\
        FedPAC  & 2.85$\times$ & Data Rep. + 1.00$\times$ &   1.01$\times$ \\
        FedAli    & 1.09$\times$ & 2.38$\times$ & 1.84$\times$ \\
        \bottomrule
    \end{tabular}
    }

    \label{tab:comparison}
\end{table}

\begin{figure}[H]
\centering
\includegraphics[width=0.5\linewidth]{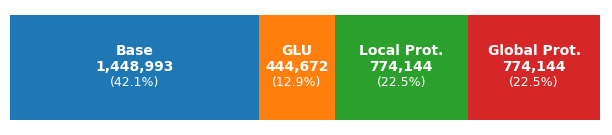}
\caption{Visualization of parameter distribution among the Base model, GLU, Local Prototype, and Global Prototype components }
\label{Fig:model parameter}
\end{figure}

Figure \ref{Fig:model parameter} illustrates the parameter count for each newly introduced component of the ALP layer alongside the small HART model. Indeed, the inclusion of local and global prototypes in each encoder block results in a significant increase in parameter size. However, it is important to note that these prototypes are non-trainable parameters and are unaffected during the back-propagation calculation. Moreover, during inference, only the local prototypes are used, ensuring that the model remains relatively efficient despite the increase in overall size.

Beyond component analysis, we also examined the impact of different distance metrics for computing the similarity/cost matrix in our framework. As shown in the second part of Table \ref{tab:ablation}, we explored alternatives such as cosine similarity and Euclidean distance, but our results indicate that the iterative Sinkhorn-Knopp technique, which operates within the Wasserstein space, achieves the most balanced performance. This advantage can be attributed to the distribution-to-distribution matching property of optimal transport methods, which contrasts with conventional point-to-point approaches. Using this characteristic, our method enables the use of prototypes in a uniform way.

\begin{table}[ht]
    \centering
    \caption{Ablation study on the HHAR dataset trained with 200 CR}
    \resizebox{\linewidth}{!}{
    \begin{tabular}{lccc}
        \toprule
        \textbf{Method} & \textbf{Personalization} & \textbf{Generalization} & \textbf{Global} \\
        \midrule
        \textbf{Configurations} & & & \\
        FedAli & 97.42 ± 3.30 & \textbf{70.04 ± 5.01} & \uline{82.80 ± 0.47} \\
        w/o GLU & \uline{97.55 ± 3.40} & 67.58 ± 5.31 & 81.22 ± 0.44 \\
        w/o Global Prototype & \textbf{97.60} ± 3.30 & 68.58 ± 4.97 & 81.94 ± 0.73 \\
        w/o Local Prototype & 97.22 ± 3.86 & \uline{68.73 ± 5.43} & \textbf{82.99 ± 0.73} \\
        FedAvg	& 96.30 ± 4.68 & 64.14 ± 4.82	& 77.86 ± 0.58 \\
        \midrule
        \textbf{Distances} & & & \\
        Wasserstein & \uline{97.42 ± 3.30} & \textbf{70.04 ± 5.01} & \textbf{82.80 ± 0.47} \\
        Cosine & \textbf{97.61} ± 3.36 & 69.08 ± 4.82 & 81.98 ± 0.52 \\
        Euclidean & 97.12 ± 3.38 & \uline{69.12 ± 5.02} & \uline{82.01 ± 0.44} \\
        \bottomrule
    \end{tabular}
    }
    \label{tab:ablation}
\end{table}

\subsection{Sensitivity analysis}
\label{sec:sensitivity}

The parameter $\beta$ controls the influence of the prototypes on their assigned embeddings within each ALP layer. To determine its optimal value, we performed tuning in five settings: 0.1, 0.2, 0.3, 0.4, and 0.5, averaging results over three random runs on HHAR clients with 200 communication rounds of training (Table \ref{tab:coefficients}). Our findings indicate that lower $\beta$ values reduce the regularization effect of global prototypes, leading to greater personalization but lower generalization and global scores. In contrast, larger values $\beta$ improve generalization and global scores at the cost of personalization. In general, we find $\beta = 0.2$ to provide the best balance for our requirements.
\begin{table}[H]
    \centering
    \caption{FedAli with different prototype influence coefficients ($\beta$)  trained with 200 CR}

    \begin{tabular}{cccc}
        \toprule
        \textbf{$\beta$} & \textbf{Personalization} & \textbf{Generalization} & \textbf{Global} \\
        \midrule
        0.1 & \textbf{97.63 ± 3.35} & 69.03 ± 5.18 & 82.27 ± 0.77 \\
        0.2 & \uline{97.42 ± 3.30} & \uline{70.04 ± 5.01} & 82.80 ± 0.47 \\
        0.3 & 97.37 ± 3.25 & 70.00 ± 4.61 & \uline{83.00 ± 0.43} \\
        0.4 & 97.13 ± 3.23 & 69.97 ± 4.60 & 82.88 ± 0.47 \\
        0.5 & 96.99 ± 3.28 & \textbf{70.30 ± 4.87} & \textbf{83.00 ± 0.31} \\
        \bottomrule
    \end{tabular}
    
    \label{tab:coefficients}
\end{table}

Next, we examine the impact of different prototype counts. Specifically, we tested three configurations: (1024, 512, 256, 128, 64, 32), (2048, 1024, 512, 256, 128, 64) and (4096, 2048, 1024, 512, 256, 128), where larger prototype counts are assigned to layers closer to the input. As shown in Table \ref{tab:protCount}, our results indicate that the performance differences between configurations are marginal. However, the setting with (2048, 1024, 512, 256, 128, 64) provided a slightly improved overall balance in performances.
\begin{table}[H]
    \centering
    \caption{FedAli with different number of prototypes across 6 layers trained with 200 CR}
    \resizebox{\linewidth}{!}{
    \begin{tabular}{ccccc}
        \toprule
        \textbf{Prototype Count} & \textbf{Per.} & \textbf{Gen.} & \textbf{Global} & \textbf{Param.}\\
        \midrule
        $1024,512,256,128,64,32$ & \textbf{97.48 ± 3.34} & \uline{69.89 ± 4.81} & \uline{82.80 ± 0.66} & 2,277,662\\
        $2048,1024,512,256,128,64$ & \uline{97.42 ± 3.30} & \textbf{70.04 ± 5.01} & \textbf{82.80 ± 0.47} & 2,664,734\\
        $4096,2048,1024,512,256,128$ & 97.38 ± 3.34 & 69.51 ± 4.82 & 82.62 ± 0.57 & \textbf{3,438,878}\\

        \bottomrule
    \end{tabular}
    }
    \label{tab:protCount}
\end{table}

Although experiments here show that an additional number of prototypes leads to small improvement in generalization, it also results in a higher parameter count. For instance, the largest configuration introduces over 3.4M parameters, whereas our selected configuration maintains a more balanced trade-off with 2.66M parameters. However, we must take into account that very heterogeneous environments with many clients, such as those represented by the RealWorld and Combined datasets, would potentially further benefit from the increased number of prototypes to better represent the large diversity.

\section{Conclusion}


In this paper, we propose FedAli, a novel PFL framework designed to enhance the robustness of mobile client models and mitigate the challenges posed by data heterogeneity. At the core of FedAli is the ALP layer, which leverages prototype-based alignment to dynamically influence incoming embeddings in both model training and inference.  Through extensive experiments in heterogeneous distributed environments, we demonstrated that FedAli significantly outperforms existing baselines in generalization while maintaining strong personalization capabilities.


\bibliographystyle{ACM-Reference-Format}
\bibliography{sample-base}


\end{document}